\documentclass[11pt]{article}
\usepackage{rotating}
\usepackage{acl2015}
\usepackage{times}
\usepackage{url}
\usepackage{latexsym}
\usepackage{tabularx}
\usepackage{makecell}
\usepackage{hyperref}
\usepackage{multirow}
\usepackage{multicol}
\usepackage{booktabs}
\usepackage{CJK}
\usepackage[section]{placeins}
\usepackage[hang,flushmargin]{footmisc} 
\usepackage[T1]{fontenc}
\usepackage{lipsum}
\newcommand\blfootnote[1]{%
  \begingroup
  \renewcommand\thefootnote{}\footnote{#1}%
  \addtocounter{footnote}{-1}%
  \endgroup
}
\usepackage{fancyhdr}
\title{Towards Better Instruction Following Language Models for Chinese: \\ Investigating the Impact of Training Data and Evaluation}

\author{Yunjie Ji, Yan Gong, Yong Deng, Yiping Peng, Qiang Niu, Baochang Ma{*}, Xiangang Li \\
Beike Inc., Beijing, China  \\
\texttt{\{jiyunjie001,gongyan013,dengyong013,pengyiping001,}  \\
  \texttt{niuqiang002,mabaochang001,lixiangang002\}@ke.com}}

\begin{document}
\maketitle
\begin{abstract}
Recently, significant public efforts have been directed towards developing low-cost models with capabilities akin to ChatGPT, thereby fostering the growth of open-source conversational models. 
However, there remains a scarcity of comprehensive and in-depth evaluations of these models' performance.
In this study, we examine the influence of training data factors, including quantity, quality, and linguistic distribution, on model performance. 
Our analysis is grounded in several publicly accessible, high-quality instruction datasets, as well as our own Chinese multi-turn conversations.
We assess various models using a evaluation set of 1,000 samples, encompassing nine real-world scenarios. 
Our goal is to supplement manual evaluations with quantitative analyses, offering valuable insights for the continued advancement of open-source chat models.
Furthermore, to enhance the performance and training/inference efficiency of models in the Chinese domain, we extend the vocabulary of LLaMA – the model with the closest open-source performance to proprietary language models like GPT-3 – and conduct secondary pre-training on 3.4B Chinese words.
We make our model, data, as well as code publicly available\textsuperscript{1}\blfootnote{
    \textsuperscript{*}Corresponding author \\
\textsuperscript{1}https://github.com/LianjiaTech/BELLE
}.
\end{abstract}

\section{Introduction}
Large language models exhibit remarkable comprehension, generation, and reasoning capabilities that surpass those of their smaller counterparts\cite{LanguageModelsAre2020,OPTOpenPretrained2022,PaLMScalingLanguage2022,BLOOM176BParameterOpenAccess2022,GPTNeoX20BOpenSourceAutoregressive2022,TrainingComputeOptimalLarge2022}. 
By learning from high-quality human-generated data, these models align with human behavior and respond to human instructions, rendering them accessible to non-experts in Natural Language Processing for various applications\cite{TrainingLanguageModels2022,ConstitutionalAIHarmlessness2022,FineTuningLanguageModels2020,LearningSummarizeHuman2022,RedTeamingLanguage,WebGPTBrowserassistedQuestionanswering2022,PretrainingLanguageModels2023}. 
Nonetheless, the development of such expansive models has become increasingly exclusive, with data, models, and algorithms no longer publicly accessible. 

To address this issue, the open-source community has undertaken\cite{ji2023exploring2,alpaca,peng2023instruction,vicuna2023,koala_blogpost_2023,xu2023baize,chinese-llama-alpaca,Dolly,Dolly2.0} numerous effective endeavors, discovering that satisfactory instruction-following performance can be achieved using smaller models with only tens of thousands of data points. 
The majority of these efforts are based on LLaMA\cite{touvron2023llama}, a series of open-source models that yield results comparable to proprietary models like GPT-3 and Gopher but with fewer parameters. 
GPT-3.5\cite{gpt3.5} and GPT-4\cite{gpt4} have also been employed to generate high-quality aligned data.
Stanford Alpaca\cite{alpaca} utilized the self-instruct method to generate 52,000 samples using GPT-3.5 based on 175 seed tasks, while LLaMA-GPT4\cite{peng2023instruction} leveraged GPT-4 for higher-quality responses.
Vicunna\cite{vicuna2023} trained LLaMA on 70,000 real user conversations with ChatGPT, and Baize\cite{xu2023baize} enabled ChatGPT to generate multi-turn conversations. 
Table \ref{overview} details the training and evaluation methods for these open-source initiatives. 
Although these projects significantly advance the development of open-source chatbots, there remains a dearth of in-depth evaluation and comparison of these models. 
Current assessments are either insufficient in quantity or solely reliant on human evaluation. 
It is imperative to investigate how these high-quality datasets can enhance instruction-following models.

In this paper, we investigate the influence of various factors, such as the quantity, quality, and linguistic distribution of training data on model performance. 
Our evaluation dataset consists of  1,000 Chinese instruction samples spanning nine real user scenarios. 
It is worth noting that our evaluation set is still not comprehensive enough, and the scores of the model on our evaluation set may differ from the actual user experience. 
Building a diverse and high-quality evaluation set will be a long-term task to accomplish.
Furthermore, we extend LLaMA's vocabulary and pre-trained it on Chinese corpora to boost its efficiency in processing Chinese data, resulting in a reduction of 60\% training and inference time without compromising performance.

\section{Related work}

\begin{table*}[h]
\centering
\caption{A simple overview of  public available chat models. More details could be found in \ref{rw_pacm}. }
\begin{tabular}{m{1.2cm}|m{1.2cm}|m{1.7cm}|m{3.3cm}|m{3cm}|m{3.2cm}}
\hline
\textbf{Project} & \textbf{Base model} & \textbf{Training} & \textbf{Training data} & \textbf{Evaluation data} & \textbf{Evaluation method}\\
\hline
Stanford alpaca  & LLaMA & Full-parameter finetuning & 52K text-davinci-003  generated instruction  data & 252 samples from self-instruct evaluation dataset & Human evaluation \\
\hline
LLaMA-GPT4 & LLaMA & Full-parameter finetuning & 52K GPT-4  generated instruction data &  \makecell[l]{1. 252 user-oriented \\ instructions\\  2. 80 vicuna test \\ samples} & \makecell[l]{1. Human evaluation \\ 2. Automatic GPT-4 \\evaluation} \\
\hline
Vicuna & LLaMA & Full-parameter finetuning & 70K user-shared conversations with ChatGPT & 80 vicuna test samples & \makecell[l]{Automatic GPT-4 \\evaluation} \\
\hline 
Koala & LLaMA & Full-parameter finetuning & \makecell[l]{ 1. Stanford alpaca \\2. Anthropic HH \\3. OpenAI webgpt \\4. OpenAI summariz-\\ation } & \makecell[l]{1. 180 samples from \\ self-instruct evalu-\\ ation dataset \\ 2. 180 real user que-\\ries that were posted \\online} & Human evaluation \\
\hline
Dolly & GPT-J & Full-parameter finetuning & Stanford alpaca & - & Case demonsrtration \\
\hline
Dolly 2.0 & Pythia & Full-parameter finetuning & 15k human-written instruction data & - & Case demonsrtration \\
\hline
Baize & LLaMA & LoRA & 15K ChatGPT generated multi-turn conversations & - & Case demonsrtration \\
\hline
\end{tabular}
\label{overview}
\end{table*}

\subsection{Large language models}
Transformer-based language models, especially the generative large language models have greatly advanced the development of Natural Language Processing \cite{vaswani2017attention,devlin2018bert,lan2019albert,yang2019xlnet,dong2019unified,clark2020electra,raffel2020exploring,LanguageModelsAre2020,OPTOpenPretrained2022,PaLMScalingLanguage2022,GPTNeoX20BOpenSourceAutoregressive2022,TrainingComputeOptimalLarge2022,glaese2022improving,srivastava2022beyond}.
The GPT (Generative Pre-trained Transformer) family of models is a remarkable instance, and its ability to comprehend and adhere to human instructions has been enhanced by RLHF \cite{TrainingLanguageModels2022,ConstitutionalAIHarmlessness2022,FineTuningLanguageModels2020,LearningSummarizeHuman2022,RedTeamingLanguage,WebGPTBrowserassistedQuestionanswering2022,PretrainingLanguageModels2023} in ChatGPT. 
As a result, ChatGPT has evolved from being a basic NLP task solver to a complete natural language assistant that can perform duties such as generating conversations and detecting errors in a piece of code.

\subsection{Instruction tuning}
Instruction-tuning is a new trend emerging from \cite{FinetunedLanguageModels2021,MultitaskPromptedTraining2021,mishra2021cross}, which seeks to improve the performance of language models by teaching them to follow natural language.
By formatting all tasks into natural language, generative language models are capable of dealing with almost all of NLP tasks. 
Early research focused on instruction tuning a general NLP task solver, and there is a trend towards converting more and more NLP datasets into a unified dataset then conducting multi-task training \cite{xu2022zeroprompt,xie2022unifiedskg,wang2022super,khashabi2020unifiedqa,min2021metaicl,ye2021crossfit,liu2019multi,zhong2021adapting,ScalingInstructionFinetunedLanguage2022}.
However these models still struggle with understanding general human instructions especially in real-world use cases.
Until the emergence of training methods like RLHF \cite{TrainingLanguageModels2022,ConstitutionalAIHarmlessness2022,FineTuningLanguageModels2020,LearningSummarizeHuman2022},  models truly began to understand various human instructions and produce good responses.

\subsection{Public available chat models}
\label{rw_pacm}
Recently, there are many attempts toward public available models.
\newcite{alpaca} trained a model on 52K instruction-following samples generated in the style of self-instruct \cite{wang2022self2} using text-davinci-003, then they did a blind pairwise comparison between Alpaca and text-davinci-003 on 252 evaluation samples from the self-instruct evaluation set. 
Instead, based on the same seed tasks, \newcite{peng2023instruction} generated 52K instruction-following samples using GPT-4. They assess their models on 252 user-oriented instructions by human based on the HHH criteria \cite{GeneralLanguageAssistant2021}.
\newcite{vicuna2023} trained a chat model based on LLaMA with a dataset consisting of 70K user-shared conversations with ChatGPT \cite{ShareGPT}.
They conducted a automatic evaluation with GPT-4 on 80 evaluation examples.
\newcite{koala_blogpost_2023} not only used ChatGPT generated datasets, but also used open source human-written instruction-following data. 
\newcite{Dolly} finetuned GPT-J with the dataset from Stanford Alpaca.
To obtain a fully and truly open instruction-tuned LLM, \newcite{Dolly2.0} established a dataset of 15K human-generated instruction data and trained a model based on Pythia. 
Both of their model and dataset are licensed for research and commercial use.
\newcite{xu2023baize} focused on  training a model capable for multi-turn dialogue  in a low-resource setting.
Therefore, they leveraged ChatGPT to engage in a conversation with itself, simulating both user and AI responses. Meanwhile, they finetuned LLaMA with a parameter-efficient tuning approach\cite{hu2021lora}.

\subsection{Evaluation of LLMs}
There are many evaluations of large language models, such as OPT \cite{OPTOpenPretrained2022}, BLOOM \cite{BLOOM176BParameterOpenAccess2022}, GLM \cite{zeng2023glm-130b}, and GPT-3 \cite{LanguageModelsAre2020}, in various tasks. \cite{HolisticEvaluationLanguage2022} conducted a thorough evaluation of 30 large language models.
\cite{ChatGPTGeneralPurposeNatural2023} evaluated the performance of ChatGPT on various NLP tasks. 
\cite{ye2023comprehensive} compared the capabilities of GPT and GPT-3.5 series models. 
\cite{bang2023multitask} compared the reasoning, hallucination reduction, and interactivity abilities of ChatGPT in multiple languages and modalities.
However many evaluation data consist of  traditional NLP tasks, which differ from real-world human usage scenarios.
\cite{srivastava2022beyond} provided 204 tasks, which are believed to be beyond the capabilities of current large language models. 
\cite{suzgun2022challenging} selected the 23 most difficult tasks from BIG-Bench, forming BIG-Bench Hard (BBH).

\section{Collect Conversation data}
\label{collect_clean_data}
ChatGPT has demonstrated a remarkable ability to generate instruction-following data, which enables the model to achieve strong performance on various tasks. 
However, its ability to engage in multi-turn conversations still falls short due to its limited contextual understanding based solely on this type of data. 
Thus, we further task ChatGPT to generate multi-turn conversation data, where it needs to generate dialogue between users and AI assistants across multiple turns.
As ChatGPT tends to produce limited and repetitive dialogue scenarios such as weather queries and airplane ticket reservations, we prompt ChatGPT with the first round of a conversation to determine the topic of the conversation and then let ChatGPT continue the conversation accordingly.

\begin{figure} [t!]
	\centering
	\includegraphics[scale=0.4]{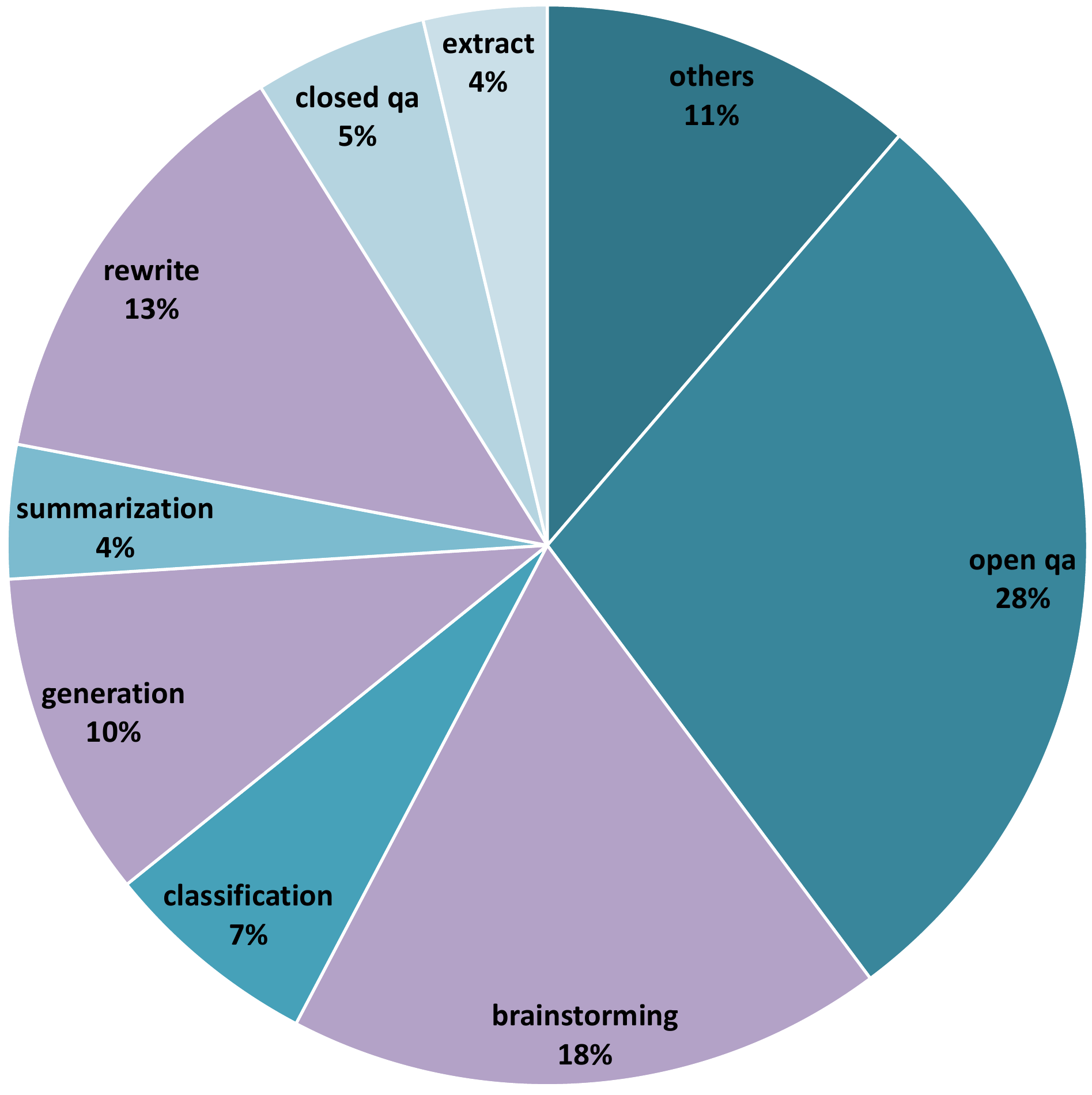}
	\caption{Task category distribution of our evaluation set. 
 We classify math and code tasks into ``others''.
 For rewrite, generation and brainstorming, no gold response is provided. As a consequence, ChatGPT is asked to evaluate these tasks without referring to  gold responses, as we believe there are more than one reasonable responses for each instruction of these three task categories. }
\label{cate_distri}
\end{figure}

\begin{figure} [t!]
	\centering
	\includegraphics[scale=0.62]{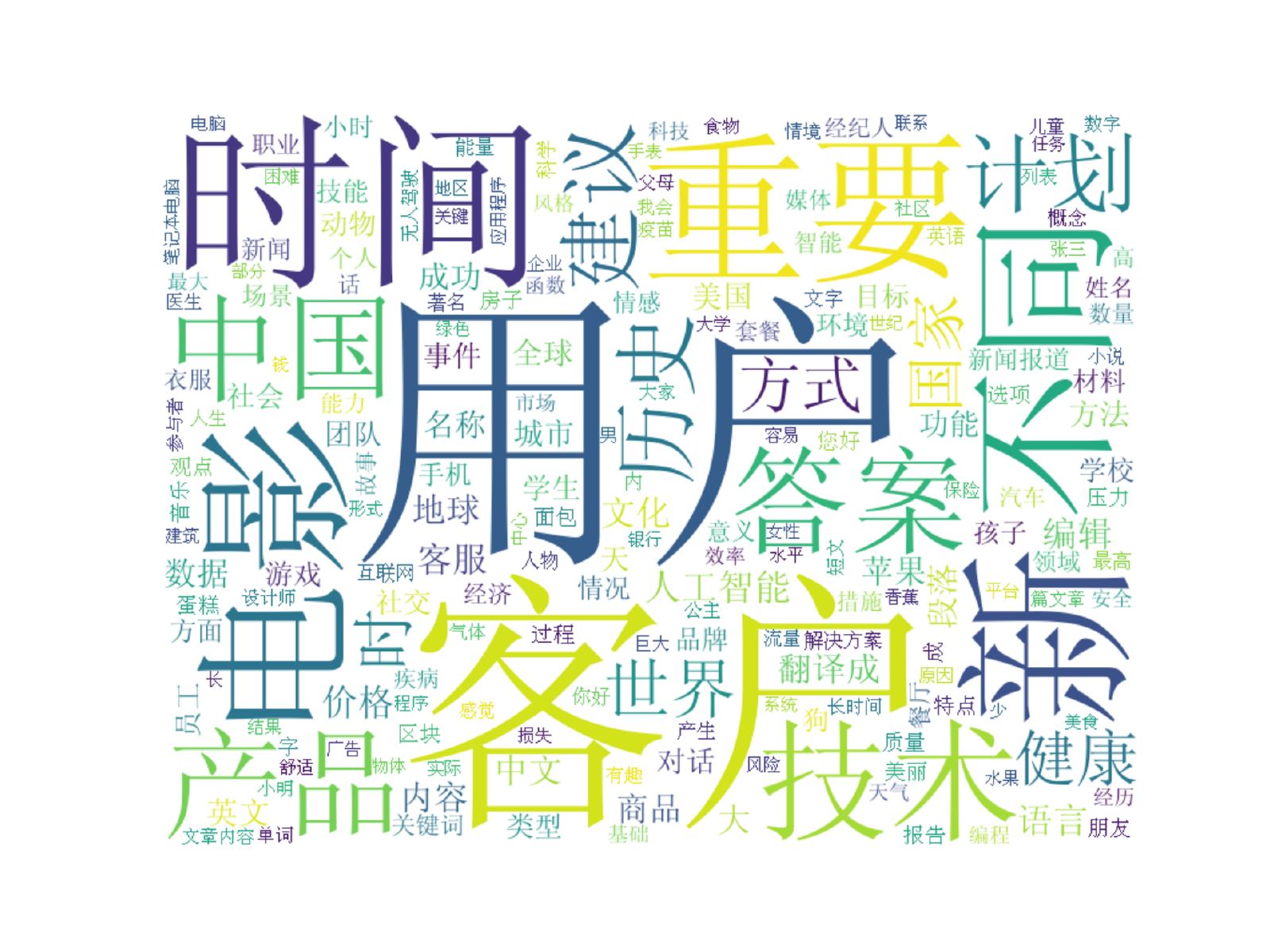}
	\caption{Word cloud of the evaluation set.}
\label{word_cloud}
\end{figure}

\noindent \textbf{Data Cleaning.} 
While ChatGPT can generate high-quality data at a relatively low cost, the generated data still suffers from issues such as repetition and logical inconsistency. 
To improve the quality of the generated data, we first remove duplicates at both token-level and semantic-level. 
Next, we use metrics such as perplexity (PPL) to select high-quality data and ensure diversity in the data by examining word frequency distributions.

\section{Evaluation data}
\begin{figure*} [t!]
	\centering
	\includegraphics[scale=0.5]{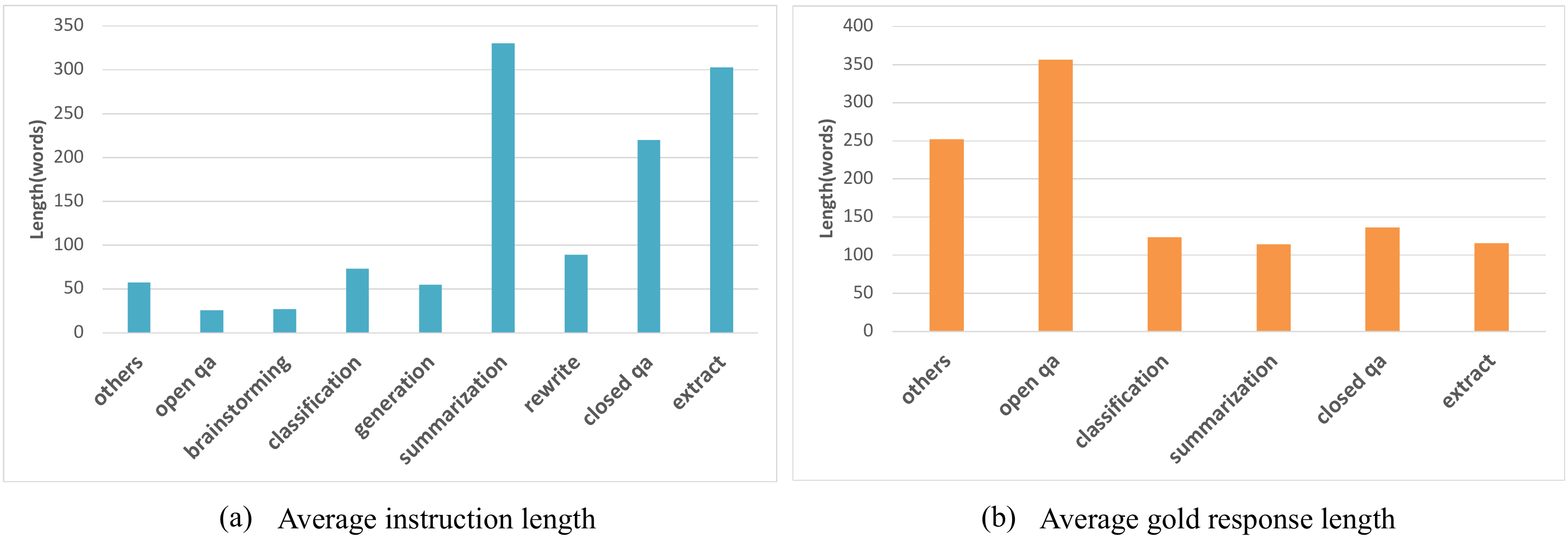}
	\caption{(a) shows average length of instructions, (b) show average length of gold responses.}
\label{length}
\end{figure*}

The evaluation data used in this paper is refined from \newcite{ji2023exploring2}.
We deduplicate the original evaluation data semantically and re-classify math and code tasks into others.
There are two reasons: firstly, these tasks are relatively difficult and current open-source models do not perform well on them, which could affect our evaluation of other abilities, secondly, ChatGPT is not reliable enough in evaluating these two tasks, which could lead to biased experimental results.
Figure \ref{cate_distri} depicts the task category distribution, which is not balanced, so we use macro-averaging when calculating overall scores.
Figure \ref{word_cloud} shows the word cloud of the evaluation set. We find that ChatGPT tends to generate data on certain specific topics.
Figure \ref{length} shows the length of evaluation samples.

\section{Extend vocabulary}
Due to the lack of optimization for Chinese language in LLaMA's vocabulary construction, a Chinese character may be split into 2 to 3 byte tokens, which severely affects the model's fine-tuning and inference speed on Chinese data\cite{chinese-llama-alpaca}. 
In order to address this issue, we train a tokenizer based on the byte-pair encoding (BPE) algorithm using sentencepiece\cite{kudo2018sentencepiece} on 12M lines of Chinese text, and set its vocabulary size to 50K. 
We merge the trained new vocabulary with the original LLaMA vocabulary, resulting in a new vocabulary of 79,458 tokens.
After that, we resize word embeddings and further pretrain LLaMA on 3.4B Chinese words with other parameters fixed. 
We test the extended tokenizer and the original tokenizer on 5,000 lines of Chinese text, and the average tokens of a line reduces from 733 to 291.

\section{Experiments}

\begin{table*}[h]
\centering
\caption{This table shows performances of models instruction-tuned with different base models and training data. We divide experimental results according to training data factors which may have great impact on model performance. \textbf{Score\_w/0\_others} is macro-averaged on categories excluding math and code. Please refer to Appendix a for scores on each category and averaged on all task categories. }  
\begin{tabular}{ccccc}
 \hline
        \textbf{Factor} & \textbf{\makecell[c]{Base model}} & \textbf{\makecell[c]{Training data}} &  \textbf{\makecell[c]{Score\_w/o\_others}} \\ 
        \hline
         \multirow{2}{*}{\makecell[c]{Extending vocabulary}} & \makecell[c]{LLaMA-EXT} & \makecell[c]{zh(alpaca-3.5\&4) +  sharegpt}     & \textbf{0.670}  \\ 
         & \makecell[c]{LLaMA} & \makecell[c]{zh(alpaca-3.5\&4) +  sharegpt}      &0.652  \\
        \hline
        \multirow{2}{*}{\makecell[c]{Data quality}} & \multirow{2}{*}{\makecell[c]{LLaMA-EXT}} & zh(alpaca-3.5)& 0.642   \\ 
         &  & zh(alpaca-4)&  \textbf{0.693}  \\ 
         \hline
         \multirow{4}{*}{\makecell[c]{Data linguistic distribution}} & \multirow{4}{*}{\makecell[c]{LLaMA-EXT}} & zh(alpaca-3.5\&4)&\textbf{0.679}    \\ 
          &  & en(alpaca-3.5\&4)  & 0.659  \\ 
          &  & \makecell[c]{zh(alpaca-3.5\&4) +  sharegpt}   & 0.670  \\ 
          &  & \makecell[c]{en(alpaca-3.5\&4) +  sharegpt}    &0.668   \\
         \hline
         \multirow{2}{*}{\makecell[c]{Data quantity}} & \multirow{2}{*}{\makecell[c]{LLaMA-EXT}} & \makecell[c]{zh(alpaca-3.5\&4) +  sharegpt}    & 0.670  \\ 
         &  & \makecell[c]{zh(alpaca-3.5\&4) +  sharegpt  + belle-3.5}  &\textbf{0.762}    \\
         \specialrule{\heavyrulewidth}{\aboverulesep}{0pt}
         \multirow{1}{*}{\makecell[c]{-}} & \makecell[c]{ChatGPT} & -    & 0.824   \\ 
         \hline
\end{tabular}
\label{main_results}
\end{table*}

\subsection{Base model}
In our experiments, two base models with 7B parameters are adopted:

\textbf{LLaMA}\cite{touvron2023llama}, which is released by Meta AI. 

\textbf{LLaMA-EXT}, which is obtained by extending the vocabulary of the vanilla LLaMA and further pre-train on 3.4B Chinese words in which only  word embeddings are updated. 

\subsection{Training settings}
All of the instruction-following models are finetuned with the same hyper-parameter settings as the open source projects with the exception of a smaller learning rate. 
No dev set is used while training models, and the last checkpoint is adopted for evaluation. 
Table \ref{hyper-parameters} lists the hyper parameters. 
We conduct experiments on 8 A100 GPUs, each has 80G memory. 

\begin{table}[t!]
\caption{Hyper-parameter settintgs, which are the same as previous open-released chat model with the exception of a smaller learning rate. }
\begin{center}
\begin{tabular}{l|r} 
\hline 
\textbf{Hyper parameter} & \textbf{Value} \\
\hline   
Precision  & bf16 \\
\hline
Epochs  & 3 \\
\hline
Batch size  & 32 \\
\hline
Learning rate  & 5e-6 \\
\hline
Weight decay  & 0 \\
\hline
Warmup ratio  & 0.03 \\
\hline
LR scheduler type  & cosine \\
\hline
Max length  & 2048 \\
\hline
\end{tabular}
\end{center}
\label{hyper-parameters}
\end{table}

\subsection{Dataset}
Our experiments are conducted using six datasets, of which five are publicly available and one is proprietary. 

\textbf{Alpaca-3.5-en}\cite{alpaca}, which is released by Stanford Alpaca and consists of 52K instruction-following samples. These samples are generated by text-davinci-003. 

\textbf{Alpaca-3.5-zh}\cite{chinese-llama-alpaca}, which is the translated Chinese version of alpaca-3.5-en.

\textbf{Alpaca-4-en, Alpaca-4-zh}\cite{peng2023instruction}, which are released by  LLaMA-GPT4, both containing 52K instruction -following samples. 
These samples are generated by GPT-4.
To obtain alpaca-4-zh, \newcite{peng2023instruction} first used ChatGPT to translate 52K instructions into Chinese then asked GPT-4 to answer them in Chinese.

\textbf{ShareGPT}\cite{ShareGPT}, which are user-shared conversations with ChatGPT, consisting of 8.3K samples. 
We conduct three steps of data cleaning \cite{vicuna2023}. 
Only English and Chinese conversations are kept. 
Besides, conversations are divided into smaller segments with a maximum length of 2048 tokens.
Finally, we derive 120,009 conversations.

\textbf{Belle-3.5}, which is our own dataset, consisting of instruction-following samples and multi-turn conversations. 
This dataset contains 500,000 samples that are filtered out from 2.3M raw data with the cleaning method mentioned in section \ref{collect_clean_data}. 

To simplify dataset names while conducting experiments under different data settings, we define two functions for identifying the language of the given dataset. \texttt{zh(d)} means the Chinese version of \texttt{d} and \texttt{en(d)} means the English one. 

\subsection{Metric}
ChatGPT is asked to evaluate responses generated by instruction-following models.
For all instructions, ChatGPT gives a score between 0 and 1, where score 0 is the worst and score 1 is the best. 
In order to reduce randomness, we set the temperature to 0.001 for model generation. 
Evaluation is achieved by invoking gpt-3.5-turbo API at the time of April 15, 2023.

We calculate model's scores for each task category and derive its overall performance on the evaluation set using macro average across these categories. 
Given ChatGPT's limitations in evaluating mathematical and coding tasks, we separately compute the scores that include all categories (denoted as \textbf{score}) and those that exclude these two categories (denoted as \textbf{score\_w/o\_others}).
Table \ref{main_results} presents the main experimental results. Please refer to Appendix \ref{detailed_scores} for the detailed scores on each task category.

\subsection{Main results}
\noindent \textbf{Extending vocabulary} With the same training data, finetuning on LLaMA-EXT outperforms finetuning on LLaMA, which could be due to the continued pre-training on 3.4B Chinese words, thus augmenting LLaMA's understanding of the Chinese language.

\noindent \textbf{Data quality} The performance of LLaMA-EXT, when trained with alpaca-4-zh, surpasses that of the model trained with alpaca-3.5-zh, thus highlighting the critical impact of high-quality training data on enhancing model performance.

\noindent \textbf{Linguistic distribution} Comparing the performance of LLaMA-EXT trained on zh(alpaca-3.5\&4) with the performance of LLaMA-EXT trained on en(alpaca-3.5\&4), it is observed that utilizing Chinese instruction-following data results in superior performance within Chinese testing scenarios. 
Interestingly, the disparity in performance between these two models is not substantial, which suggests that the model's capacity to comprehend instructions in one language can be effectively transferred to another language, owing to its inherent multilingual capabilities rooted from the pre-trained language model.
The scores of  zh(alpaca-3.5\&4)+sharegpt and en(alpaca-3.5\&4)+sharegpt are nearly identical.
This signifies that the addition of a modest quantity of Chinese data (which comes from sharegpt) to the English training data can dramatically enhance performance within the Chinese context.
However, it is also discovered that incorporating sharegpt into zh(alpaca-3.5\&4) does not yield any further improvements. 
This may be attributed to our inability to validate the model's proficiency in multi-turn conversations using our evaluation set.

\noindent \textbf{Data quantity} In terms of training data quantity, an increase in data quantity has been shown to significantly improve performance.
It is worth noting that such huge improvement may partly come from the similar distribution between belle-3.5 and our evaluation data. 
The categories, topics and complexity of evaluation data will brings great impact on evaluation results. 

\noindent \textbf{Compare with ChatGPT} Lastly, we identify a considerable performance gap between our model and ChatGPT. 
Considering the limited evaluation capabilities of ChatGPT and the incompleteness of our evaluation data, it is anticipated that these disparities might be even larger than the score differences observed within the evaluation set. 
This serves as motivation for us to persist in improving our model.

\subsection{Challenge of building evaluation set}
Through case analysis, we discover that our evaluation set faces limitations in its comprehensiveness, leading to an incongruity between the improvements of model scores  and actual user experience.

Constructing a high-quality evaluation set presents a significant challenge, as it necessitates the inclusion of a diverse range of usage scenarios while maintaining a balanced difficulty gradient. 
If the evaluation examples predominantly consist of difficult cases, the performance of all models will be poor, making it challenging to discern the effects of various training strategies. 
Conversely, if the evaluation examples are predominantly easy, the assessment will lose their comparative value.
Moreover, it is imperative to ascertain that the evaluation data remains distinct from the training data.

Based on these observations, we caution against the assumption that the model has achieved a performance level comparable to ChatGPT solely by obtaining favorable results on a limited number of test samples. 
We believe that prioritizing the continuous development of a comprehensive evaluation set is of great importance.

\subsection{Case demonstration}  
In Table \ref{cases}, we list several evaluation examples and corresponding model responses to qualitatively demonstrate the gap between ChatGPT and our instruction-following model. 
Our model is obtained by finetuning LLaMA-EXT with zh(alpaca-3.5\&4)+sharegpt+belle-3.5.
In our study, we find that our model tends to generate longer responses, which may be due to the presence of GPT-4 generated or multi-turn dialogue data in our training set.
In the first example, our model made an error by classifying birds as mammals, while ChatGPT classified fish as reptiles, which is also not entirely accurate.
For the following two math problems, although our model correctly identified the steps, it made mistakes in numerical calculations.
As for the final example, both models generated reasonable responses, but our model showed slight content repetition. 
This may be related to our setting of a low temperature (0.001) during generation. 
When we increase the temperature to 0.5, we observe a significant increase in diversity of responses for the same task.
\section{Conclusion}
In conclusion, this study addresses the growing need for comprehensive evaluations of open-source conversational models by investigating the influence of various training data factors, such as quantity, quality, and linguistic distribution. 
By utilizing publicly accessible high-quality instruction datasets and Chinese multi-turn conversations, we assess different models on a evaluation set of 1,000 samples across nine real-world scenarios. 
We also conclude several challenges of building a comprehensive evaluation dataset and argue the necessity of prioritizing the development of such evaluation set.
Moreover, this study extends the vocabulary of LLaMA and conducts secondary pre-training with 3.4B Chinese words to enhance its performance and efficiency in the Chinese domain. 
This results in a 60\% reduction in training and inference time without sacrificing performance. 
By making the model, data, and code publicly available, this research contributes to the ongoing efforts of the open-source community to develop more accessible and efficient conversational models especially for Chinese. 

\begin{table*}[h]
\centering
\caption{Examples of responses generated by ChatGPT and our model.}
\begin{tabular}{m{2cm}|m{8cm}|m{5cm}}
\hline
\textbf{Instruction} & \textbf{Our model} & \textbf{ChatGPT}\\
\hline
\begin{CJK*}{UTF8}{gkai}将下列五个词分为两个组别，每个组别都有一个共同点：狗、猫、鸟、鱼、蛇。
 \end{CJK*} &\begin{CJK*}{UTF8}{gkai}  将这五个词分为两个组别，可以基于它们的共性来进行分类。以下是一种可能的分类方式：
 
组别一：哺乳动物

- 狗

- 猫

- 鸟

这三个词都是哺乳动物，它们都有毛发和乳腺，能够哺育幼崽。

组别二：爬行动物

- 鱼

- 蛇

这两个词都是爬行动物，它们都有鳞片和冷血，并且通常通过产卵来繁殖后代。

这种分类方式基于这些动物的共性，即它们都是脊椎动物，并且都有特定的生理特征和行为习惯。这种分类方式也有助于我们更好地理解这些动物，以及它们与其他动物之间的相似性和差异性。 \end{CJK*} & \begin{CJK*}{UTF8}{gkai} 组别1：狗、猫、鸟
共同点：都是宠物

组别2：鱼、蛇
共同点：都属于爬行动物，没有四肢。 \end{CJK*} \\
\hline
\begin{CJK*}{UTF8}{gkai} 请给出以下数学问题的答案：$10 \div 2 \times (5-2)^2 + 7$ \end{CJK*}  &  \begin{CJK*}{UTF8}{gkai}首先，按照数学运算的优先级，需要先计算括号里的表达式，即$5-2=3$，然后将3的平方计算，即$3^2=9$。
接下来，需要先计算乘法，即$10\div2\times9=45$。最后，需要先计算除法，即$10\div2=5$，然后将乘法和除法的结果相加，即$45+5=54$。因此，$10 \div 2 \times (5-2)^2 + 7$的答案是54。\end{CJK*} & \begin{CJK*}{UTF8}{gkai}答案是 52。计算步骤如下：
1. 先计算括号内的表达式 $5-2=3$。
2. 再计算括号外的除法 $10 \div 2=5$。
3. 接着计算括号外的乘方 $3^2=9$。
4. 最后加上常数项得到结果 $5 \times 9 + 7=52$。\end{CJK*} \\
\hline
\begin{CJK*}{UTF8}{gkai} 学校买来6箱乒乓球拍，每箱8副，每副9元．一共花了多少钱？ \end{CJK*}  &  \begin{CJK*}{UTF8}{gkai} 学校买了6箱乒乓球拍，每箱8副，每副9元。为了计算总共花费的钱数，我们需要将每箱的8副乒乓球拍乘以每副的9元，然后将结果相加。
6箱 × 8副 × 9元/副 = 54元
因此，学校一共花了54元购买这些乒乓球拍。\end{CJK*} & \begin{CJK*}{UTF8}{gkai}一共买来（6×8）=48副乒乓球拍，花费的钱数为：48×9=432元。答案：432元。\end{CJK*} \\
\hline
\begin{CJK*}{UTF8}{gkai} 根据以下资料，为公司编写一个招聘广告：公司名称：ABC科技，职位：软件工程师，要求：熟练掌握Java，有2年以上工作经验。
 \end{CJK*}  &  \begin{CJK*}{UTF8}{gkai} ABC科技招聘软件工程师！
 
我们正在寻找一位熟练掌握Java，有2年以上工作经验的人才加入我们的团队。作为一家领先的技术公司，我们致力于为客户提供最优质的软件解决方案。

如果您热爱编程，对Java有着深入的理解，并且拥有2年以上相关工作经验，我们非常欢迎您加入我们的团队。

作为软件工程师，您将负责开发和维护我们的软件产品，与团队成员紧密合作，确保项目按时交付，并为客户提供最优质的技术支持。

如果您符合以上条件，请尽快联系我们！我们期待您的加入！\end{CJK*} & \begin{CJK*}{UTF8}{gkai}ABC科技是一家专注于软件开发的公司，现招聘软件工程师一名。我们要求申请者熟练掌握Java编程语言，具备至少2年以上的相关工作经验。如果你有扎实的计算机技能、良好的团队合作能力和独立解决问题的能力，欢迎加入我们！我们将向你提供有竞争力的薪酬待遇、良好的职业发展空间和良好的工作环境。如果你对此职位感兴趣，请将简历发送至我们的邮箱，并注明“软件工程师申请”。我们期待您的加入！\end{CJK*} \\
\hline
\end{tabular}
\label{cases}
\end{table*}

\bibliographystyle{acl}
\bibliography{acl2015}

\section{Appendix A}
\subsection{Detailed evaluation scores}
\label{detailed_scores}

\begin{sidewaystable*} 
  \centering 
  \caption{Detailed scores on each task category, which is complementary to \ref{main_results}. } 
  \label{tab:example} 
 \begin{tabular}{cccccccccccccc}
 \hline
        \textbf{Factor} & \textbf{\makecell[c]{Base\\ model}} & \textbf{\makecell[c]{Training\\ data}} & \textbf{others} & \textbf{rewrite} & \textbf{\makecell[c]{classif-\\ication}} & \textbf{generation} & \textbf{\makecell[c]{summari-\\zation}} & \textbf{extract} & \textbf{\makecell[c]{open\\qa}}& \textbf{\makecell[c]{brain-\\storming}} & \textbf{\makecell[c]{closed\\qa}}& \textbf{\makecell[c]{macro\\ave}} & \textbf{\makecell[c]{macro ave\_ \\ w/o\_others}} \\ 
        \hline
         \multirow{2}{*}{\makecell[c]{Extending \\vocabulary}} & \makecell[c]{LLaMA\\-EXT} & \makecell[c]{zh(alpaca-3.5\&4) \\+  sharegpt} & 0.419    & 0.858   & 0.655    & 0.897   & 0.663   & 0.456  & 0.422   & 0.837  & 0.577    & 0.643  & 0.670  \\ 
         & \makecell[c]{LLaMA} & \makecell[c]{zh(alpaca-3.5\&4) \\+  sharegpt} &0.442       & 0.828     & 0.655      & 0.853     &0.743    & 0.444    & 0.355    & 0.803    &0.537     & 0.629    &0.652  \\
        \hline
        \multirow{2}{*}{\makecell[c]{Data \\quality}} & \multirow{2}{*}{\makecell[c]{LLaMA\\-EXT}} & zh(alpaca-3.5)& 0.338  & 0.789 & 0.691  & 0.825 & 0.624  & 0.423 & 0.358  & 0.805 & 0.616  &0.608 & 0.642   \\ 
         &  & zh(alpaca-4)& 0.419  & 0.825 & 0.626  & 0.918 & 0.690  & 0.559 & 0.451 & 0.879 & 0.593  & 0.662 & 0.693  \\ 
         \hline
         \multirow{4}{*}{\makecell[c]{Data linguistic\\ distribution}} & \multirow{4}{*}{\makecell[c]{LLaMA\\-EXT}} & zh(alpaca-3.5\&4)& 0.412   & 0.807  & 0.637   & 0.889  & 0.709   & 0.489  & 0.445   & 0.814  & 0.645   &0.650  & 0.679    \\ 
          &  & en(alpaca-3.5\&4)& 0.376   & 0.760  & 0.606   & 0.894  & 0.744   & 0.489  & 0.376  & 0.899  & 0.502   & 0.627 & 0.659  \\ 
          &  & \makecell[c]{zh(alpaca-3.5\&4) \\+  sharegpt} & 0.419    & 0.858   & 0.655    & 0.897   & 0.663   & 0.456  & 0.422   & 0.837  & 0.577    & 0.643  & 0.670  \\ 
          &  & \makecell[c]{en(alpaca-3.5\&4) \\+  sharegpt} & 0.508     & 0.774    & 0.632     & 0.818    & 0.691  & 0.578   & 0.380  & 0.819   &0.653     & 0.650   &0.668   \\
         \hline
         \multirow{2}{*}{\makecell[c]{Data \\quantity}} & \multirow{2}{*}{\makecell[c]{LLaMA\\-EXT}} & \makecell[c]{zh(alpaca-3.5\&4) \\+  sharegpt} & 0.419    & 0.858   & 0.655    & 0.897   & 0.663   & 0.456  & 0.422   & 0.837  & 0.577    & 0.643  & 0.670  \\ 
         & & \makecell[c]{zh(alpaca-3.5\&4) \\+  sharegpt \\ + belle-3.5} &0.566      & 0.904    & 0.820      & 0.984    &0.753   & 0.461   & 0.564   & 0.938    &0.672    & 0.740   &0.762    \\
         \specialrule{\heavyrulewidth}{\aboverulesep}{0pt}
         \multirow{1}{*}{\makecell[c]{-}} & \makecell[c]{ChatGPT} & \makecell[c]{-} &0.875     & 0.861    & 0.813     & 0.971    & 0.795    & 0.767   & 0.690    & 0.944   & 0.751     & 0.830   & 0.824   \\ 
         \hline
\end{tabular}
\end{sidewaystable*}

\end{document}